# Dimensionality Reduction for Improving Out-of-Distribution Detection in Medical Image Segmentation

McKell Woodland[1,2], Nihil Patel[1], Mais Al Taie[1], Joshua P. Yung[1], Tucker J. Netherton[1], Ankit B. Patel[2,3], and Kristy K. Brock[1]

1. The University of Texas MD Anderson Cancer Center, Houston TX 77030, USA
mewoodland@mdanderson.org
2. Rice University, Houston TX 7005, USA
3. Baylor College of Medicine, Houston TX 77030, USA

**Abstract.** Clinically deployed segmentation models are known to fail on data outside of their training distribution. As these models perform well on most cases, it is imperative to detect out-of-distribution (OOD) images at inference to protect against automation bias. This work applies the Mahalanobis distance post hoc to the bottleneck features of a Swin UNETR model that segments the liver on T1-weighted magnetic resonance imaging. By reducing the dimensions of the bottleneck features with principal component analysis, OOD images were detected with high performance and minimal computational load.

## 1 Introduction

Deep learning (DL) models struggle to generalize to information that was not present while the model was being trained [1]. This problem is exacerbated in the medical field as collecting large-scale, annotated, and diverse training datasets is challenging due to the cost of labeling, presence of rare cases, and patient privacy. Even models that have demonstrated high-performance during external validation may fail when presented with novel information after clinical deployment. This can be demonstrated by the work of Anderson et al. [2]. On test data, 96% of their DL-based liver segmentations were deemed clinically acceptable, with the majority of their autosegmentations being preferred over manual segmentations. The two images that the model failed on contained cases that were not present during training – namely, ascites and a stent.

While autosegmentations are typically manually evaluated and corrected, if need be, by a clinician before they are used in patient treatment, the main concern is automation bias, where physicians may become too reliant on model output. Protecting against automation bias is especially important for clinically deployed segmentation models, as these segmentations influence the amount of radiation that a patient will receive during treatment. In a review study, Goddard et al. found that automation bias in healthcare can be reduced by displaying low confidence values for recommendations that are likely incorrect [3].

Displaying confidence values that correspond to the likelihood that a DL-based prediction is correct is a non-trivial problem as DL models are inherently poorly calibrated [4]. While some methods attempt to calibrate the model [5]–[8], others define an out-of-distribution (OOD) detection score [9], [10]. OOD detection operates under the assumption that the model is unlikely to perform well on data outside of the model's training distribution. While these methods perform well in theoretical settings, they often do not perform well in real-world scenarios [11]. This is especially true when these techniques are applied to medical images [12]. In fact, Cao et al. found that no method performed better than random guessing when applied to unseen medical conditions or artifacts [13].

The Mahalanobis distance is one of the most utilized OOD detection methods due to its simplicity [9]. One of the major reasons it struggles in practice is due to the curse of dimensionality. As it is a distance, it loses meaning in high-dimensional spaces and thus cannot be applied to images directly. In the classification domain, great success was achieved when the Mahalanobis distance was applied to embeddings extracted from pretrained transformers [14]. Similarly, Gonzalez et al., applied the Mahalanobis distance to embeddings extracted from an nnU-Net for medical image segmentation [15]. The major problem is that embeddings from 3D segmentation models are an order of magnitude larger than the embeddings from 2D classification models.

We build upon previous work by applying the Mahalanobis distance to principal component analysis (PCA) projected embeddings extracted from a pretrained Swin Transformer-based segmentation model. The main contributions of our paper are as follows:

1. Applying the Mahalanobis distance to a Swin UNETR model for OOD detection.
2. Reducing the dimensionality of bottleneck features using PCA before the Mahalanobis distance is applied.
3. Proposing a successful OOD detection pipeline that has minimal computation load and can be applied post hoc to any U-Net-based segmentation model.

## 2 Methods

### 2.1 Data

The training dataset was comprised of 337 T1-weighted liver magnetic resonance imaging exams (MRIs). The T1-weighted images came from the Duke Liver MRI [16], AMOS [17], [18], and CHAOS [19], [20] datasets. 27 T1-weighted liver MRIs from The University of Texas MD Anderson Cancer Center were employed for testing the segmentation model.

To protect against automation bias, OOD images should be defined as images that differ enough from the training distribution that the segmentation model is likely to fail on them. As such, the model's test data is split into in-distribution (ID) and OOD categories based on model performance. Specifically, an image is labelled ID if it has a Dice similarity coefficient (DSC) of at least 95%. Accordingly, an image is labelled OOD if it has a DSC under 95%. This follows Hendrycks et al., in the classification domain, who defined OOD data to be data that was incorrectly classified [21].

An additional 23 T1-weighted liver MRIs were acquired from The University of Texas MD Anderson Cancer Center for the OOD evaluation. All these images were flagged by physicians for poor image quality in a clinical setting. 14 images contained motion artifacts, 7 contained truncation artifacts, and the other two images contained a single artifact: magnetic susceptibility and spike noise. None had associated ground truth liver segmentations.

All test images were retrospectively acquired under an approved internal review board protocol. All images were preprocessed by reorientation to Right-Anterior-Superior (RAS), resampling to a uniform spacing (1.5, 1.5, 2.0) mm, and normalization using each image's mean and standard deviation. Example test images are shown in Figure 1.

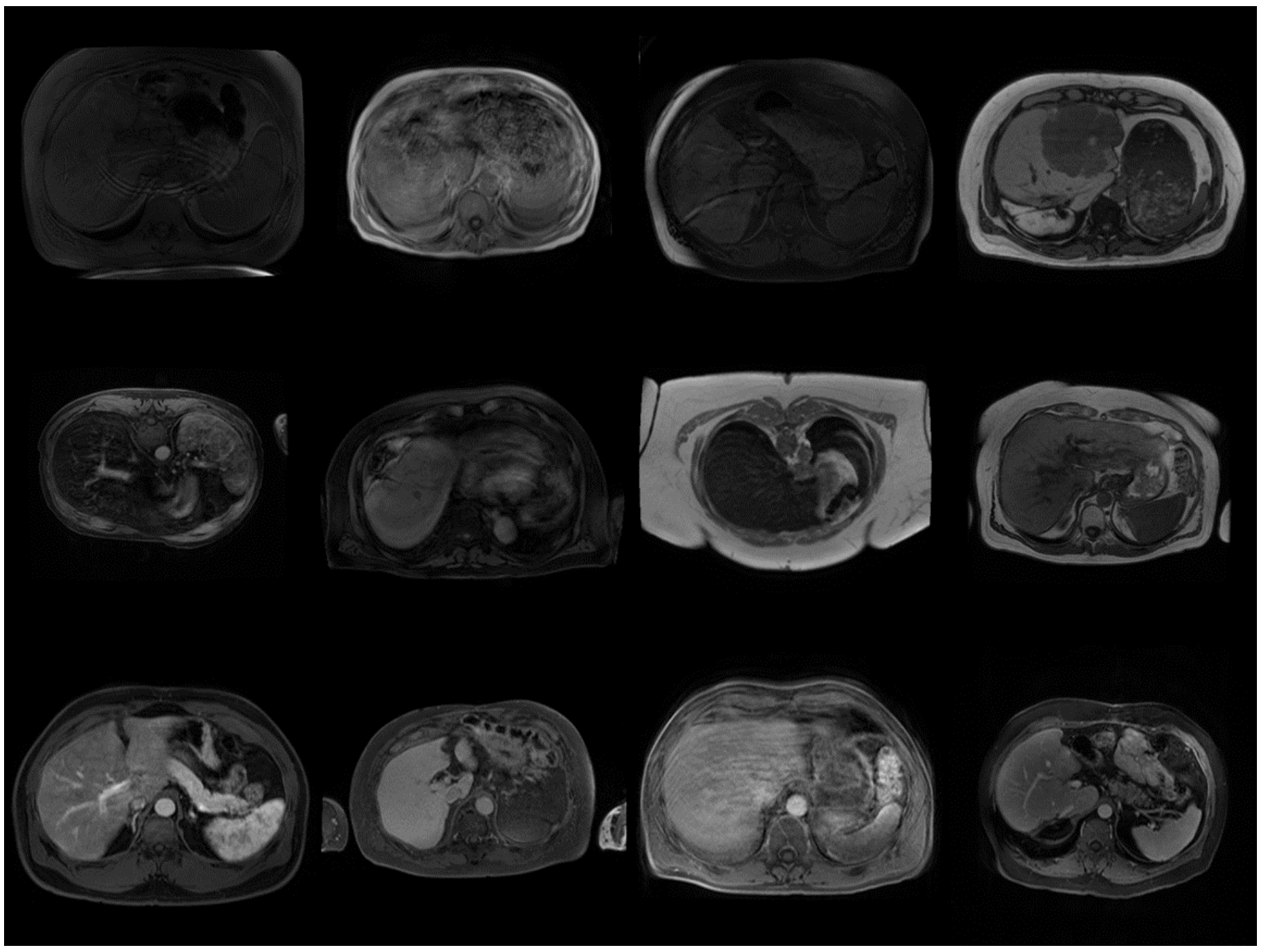

**Fig. 1.** Sample images from the test dataset. (Top) Images that were determined to be OOD by poor performance of the segmentation algorithm. (Middle) Images that were flagged for poor image quality in the clinic. (Bottom) Images that were determined to be ID by good performance of the segmentation algorithm.

## 2.2 Segmentation Model

A Swin UNETR model [22], [23] was trained to segment the T1-weighted MRIs. The encoder portion of the model was pretrained using self-distilled masked imaging (SMIT) [24] utilizing 3,610 unlabeled head and neck computed tomography scans (CTs) from the Beyond the Cranial Vault (BTCV) Segmentation Challenge dataset [25]. The official Swin UNETR codebase[1], built on top of the Medical Open Network for AI (MONAI) [26], was utilized for the pretrained weights and training. All default parameters were used, with no hyperparameter searches performed. Models were trained on a single node of a Kubernetes cluster with eight A100 graphic processing units. The final model was selected according to the weights with the highest validation DSC. It was evaluated on test images with the DSC and the Hausdorff distance.

## 2.3 Out-of-Distribution Detection

The Mahalanobis distance $D$ measures the distance between a point $x$ and a distribution with mean $\mu$ and covariance matrix $\Sigma$, $D^2 = (x-\mu)^T \Sigma^{-1} (x-\mu)$ [27]. Lee et al. first proposed using the Mahalanobis distance for OOD detection by using it to calculate the distance between test images embedded by a classifier and a Gaussian distribution fit to class-conditional embeddings of the training images [9]. Similarly, Gonzalez et al. used the Mahalanobis distance for OOD detection in segmentation networks by extracting embeddings from the encoder of a nnU-Net [15]. As distances in high dimensions are subject to the curse of dimensionality, both sets of authors decreased the dimensionality of the embeddings through average pooling. Lee et al. suggested pooling the embeddings such that the height and width dimensions are singular [9].

In our work, encoded representations of all images were extracted from the bottleneck features of the Swin UNETR models. Images were resized to (256, 128, 128) to ensure a uniform size of the encoded representations (768, 8, 4, 4). A Gaussian distribution was fit on the encodings of the training data. The Mahalanobis distance between the embedding of each test image and the Gaussian distribution was calculated. All calculations were performed on an Intel® Xeon® E5-2698 v4 @ 2.20GHz central processing unit.

As distances in extremely high-dimensional spaces often lose meaning [28], experiments were performed on the effect of decreasing the size of the bottleneck features with average pooling, principal component analysis (PCA), uniform manifold approximation and projection (UMAP) [29], and t-distributed stochastic neighbor embeddings (t-SNE) [30]. For average pooling, features were pooled in both 2- and 3-dimensions with kernel size $j$ and stride $k$ for $(j,k) \in \{(2,1),(2,2),(3,1),(3,2),(4,1)\}$. For PCA, each embedding was flattened and standardized. For both PCA and UMAP, a hyperparameter search was performed over the number of components $n$ such that $n \in \{2,4,8,16,32,64,128,256\}$. Average pooling was performed using the PyTorch Python package and PCA and t-SNE were performed using the scikit-learn Python package. UMAP was performed using the UMAP Python package [31].Outside of the hyperparameter searches mentioned above, default parameters were used.

[1] github.com/The-Veeraraghavan-Lab/SMIT

OOD detection was evaluated with the area under the receiver operating characteristic curve (AUROC), area under the precision-recall curve (AUPR), and false positive rate at 75% true positive rate (FPR75). For all calculations, OOD was considered as the positive class. As both UMAP and t-SNE are stochastic, the average was taken over 10 iterations of the algorithms. Our code can be found at github.com/mckellwoodland/dimen_reduce_mahal.

# 3 Results

The Swin UNETR achieved a mean DSC of 96% and a mean Hausdorff distance of 14 mm. 13 images had a DSC over 95% and were thus classified as ID. The remaining 14 images were classified as OOD. Figure 3 displays visual examples of the segmentation quality of the model.

The calculation of the Mahalanobis distance, as originally defined, was computationally intractable. The inverse of the covariance matrix took ~72 minutes to compute (Table 1). Once saved, it takes 75.5 GB to store the inverse. Once the matrix is in memory, it takes ~2 seconds for each Mahalanobis distance calculation. The average (±SD) Mahalanobis distance on training data was 1203.02 (±24.66); whereas, the average (±SD) Mahalanobis distance on test data was $1.47 \times 10^9$ $(\pm 8.66 \times 10^8)$ and $1.52 \times 10^9 (\pm 9.10 \times 10^8)$ for ID and OOD images respectively. The high dimensionality of the calculation resulted in poor OOD detection performance (Table 1).

Reducing the dimensionality of the embeddings not only made the Mahalanobis distance calculation more computationally feasible, but also improved the OOD detection (Table 1). While the search over average pooling hyperparameters proved to be volatile (Table S1 in the Supplementary Material), the best results were achieved with 3D convolutions that resulted in the height and width dimensions being singular, supporting the suggestion of Lee et al. [9].

**Table 1.** The AUROCs (↑), AUPRs (↑), and FPR75s (↓) for the OOD detection. ↑ means that higher is better, whereas ↓ means lower is better. Computation time is the time it takes to compute the inverse of the covariance matrix in seconds. Bold text denotes the best performance. The baseline experiment is the Mahalanobis distance calculated on the original bottleneck features. AveragePool2D(j, k) represents embeddings that were 2D average pooled with kernel size j and stride k. Similar notation applies for 3D embeddings. UMAP(n) and PCA(n) represent the respective dimensionality reduction technique being performed with n components. Only the best performing average pooling, UMAP, PCA results were included in this table. Refer to Tables S1-S3 in the Supplementary Material for the results of the full hyperparameter searches.

| Experiment | AUROC | AUPR | FPR75 | Computation Time |
|---|---|---|---|---|
| Baseline | 0.51 | 0.60 | 0.85 | 4327.4080 |
| AveragePool3D(3, 2) | 0.76 | 0.84 | 0.38 | 0.1450 |
| AveragePool3D(4, 1) | 0.70 | 0.75 | 0.31 | 0.5721 |
| UMAP(2), n=10 | 0.79 (±0.05) | 0.85 (±0.04) | 0.36 (±0.13) | 0.0002 (±0.0000) |
| t-SNE, n=10 | 0.82 (±0.05) | 0.87 (±0.04) | 0.27 (±0.14) | 0.0003 (±0.0003) |

| PCA(2) | 0.90 | 0.93 | **0.08** | 0.0001 |
|---|---|---|---|---|
| PCA(256) | **0.92** | **0.94** | 0.15 | 0.0118 |

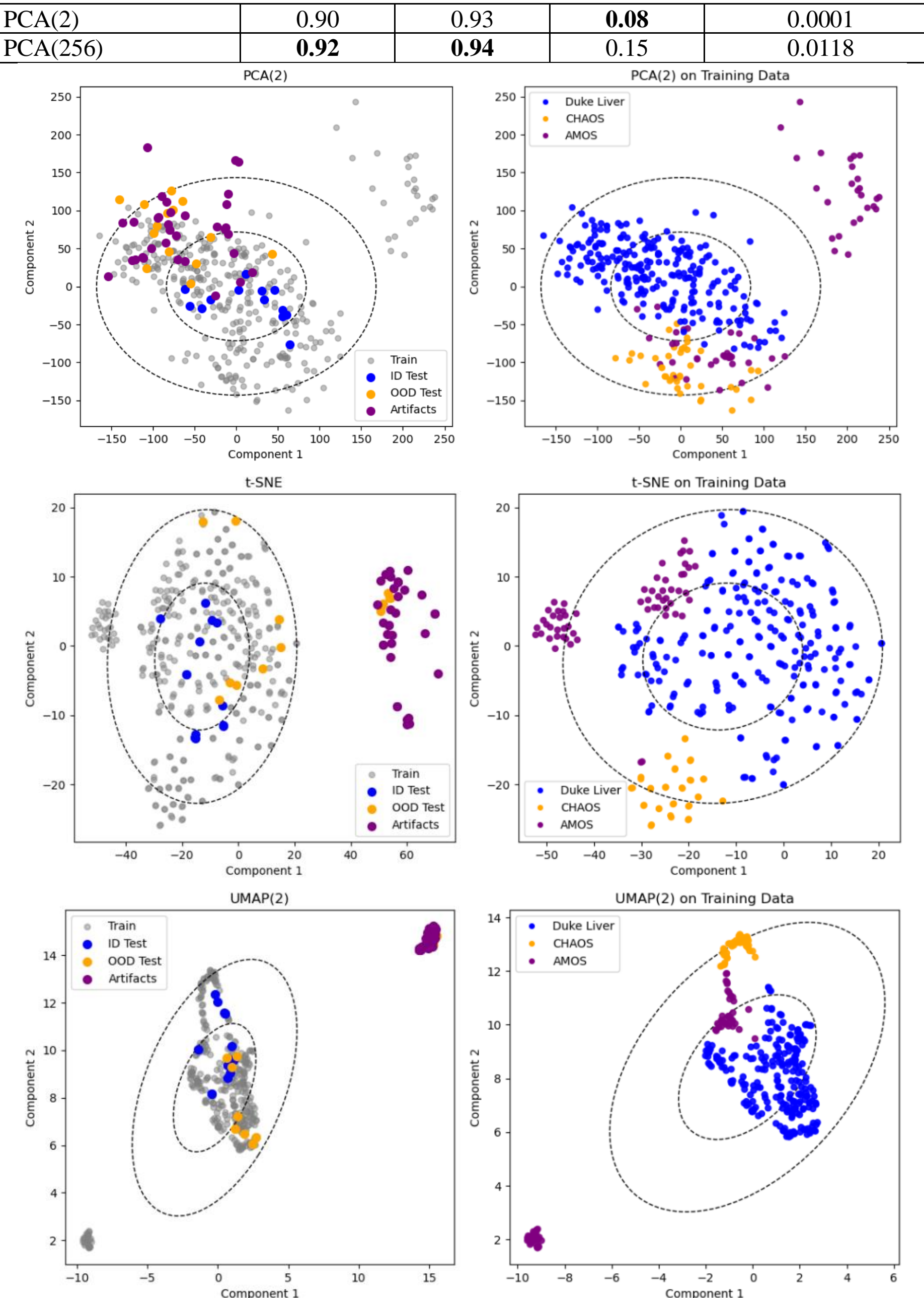

**Fig. 2.** Visualization of embeddings with two components. (Top) PCA projections. (Middle) t-SNE projections. (Bottom) UMAP projections. Projections for all data are in the left column. Projections for the training data by class are in the right column. The black ellipses are the covariance ellipses (one and two standard deviations) for the training distribution.

**Images with Low and High Mahalanobis Distances**

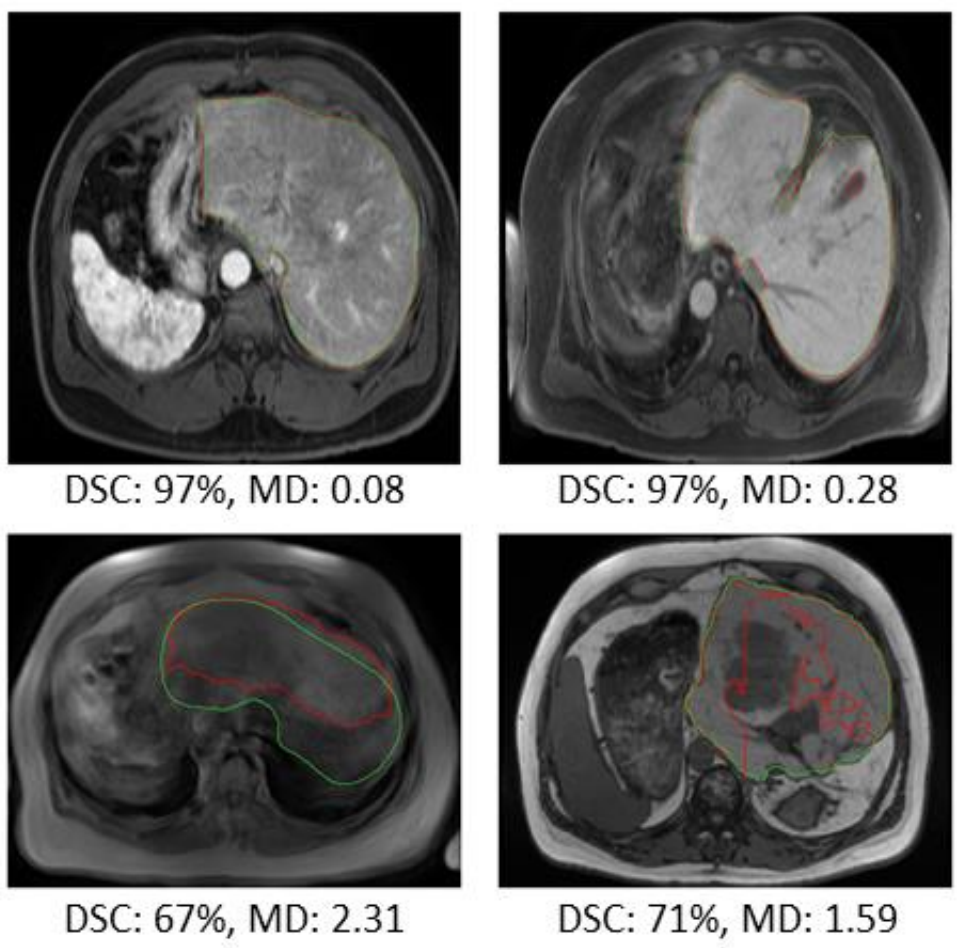

**Fig. 3.** Segmentations of images that contain low and high Mahalanobis distances (calculated on the PCA-projected embeddings with two components). Examples of images with low Mahalanobis distances are in the top row; whereas, examples with high distances are in the bottom row. Green is the ground truth segmentation; red is the automated segmentation. MD refers to the Mahalanobis distance. A higher DSC corresponds with better segmentation performance, whereas a higher distance corresponds to the image being OOD.

The best results were achieved with PCA (Table 1). Reducing the dimensionality to only two principal components was sufficient to achieve 90% AUROC, 93% AUPR, and 8% FPR75. Figure 2 demonstrates that most in-distribution test images were mapped within one standard deviation of the mean of the training distribution (the image that was not contained a motion artifact); whereas, most OOD test images were mapped between outside of the first standard deviation. The four OOD images mapped within one standard deviation had an average DSC of 88%; whereas, the OOD images mapped outside of one standard deviation had an average DSC of 79%. Additionaly,18 out of the 23 images that contained MRI artifacts were mapped outside of the first standard deviation. Furthermore, the two principal components visually cluster the different distributions within the training distribution. The 26 images from the AMOS dataset that were mapped outside of the second standard deviation were blurry. While UMAP and t-SNE did not perform as well as PCA, they still clustered the datasets in the training distribution and mapped OOD data outside of the first standard deviation. Notably, both UMAP and t-SNE mapped the data with imaging artifacts far from the training distribution. Figure 3 displays several images with the lowest and highest Mahalanobis distances for PCA with 2 components. Low distances were associated with high segmentation performance.

# 4 Conclusion

In this work, the Mahalanobis distance was applied to dimensionality-reduced bottleneck features of a Swin UNETR. The resulting pipeline was able to embed an entire 3D medical image into only two principal components. These two components were sufficient to visually cluster datasets drawn from different institutions. Additionally, only two components were required for detecting images that the segmentation model performed poorly on with high performance. In a clinical setting, a warning that the model likely failed could be added to images with large Mahalanobis distances. This would protect against automation bias, which would in turn protect patients whose scans have irregular attributes. The entire pipeline could be added post hoc to any trained segmentation model and would incur minimal computational costs.

# 5 Acknowledgments

Research reported in this publication was supported in part by the Tumor Measurement Initiative through the MD Anderson Strategic Initiative Development Program (STRIDE), the Helen Black Image Guided Fund, the Image Guided Cancer Therapy Research Program at The University of Texas MD Anderson Cancer Center, a generous gift from the Apache Corporation, and the National Cancer Institute of the National Institutes of Health under award numbers R01CA221971, P30CA016672, and R01CA235564.

## Supplementary Material

**Table S2.** The OOD detection hyperparameter search for average pooling. Bold text denotes the best performance. AveragePool2D(j, k) represents embeddings that were 2D average pooled with kernel size j and stride k. Similar notation applies for 3D embeddings. For AUROC and AUPR, higher is better. For FPR75, lower is better. Calculation time is the time it takes the calculate the inverse of the covariance matrix in the Mahalanobis distance calculation in seconds.

| Experiment | AUROC | AUPR | FPR75 | Calculation Time |
|---|---|---|---|---|
| AveragePool2D(2, 1) | 0.68 | 0.76 | 0.62 | 847.0839 |
| AveragePool2D(2, 2) | 0.66 | 0.77 | 0.69 | 135.6774 |
| AveragePool2D(3, 1) | 0.68 | 0.80 | 0.69 | 122.9558 |
| AveragePool2D(3, 2) | 0.62 | 0.71 | 0.54 | 2.2238 |
| AveragePool2D(4, 1) | 0.67 | 0.68 | 0.46 | 2.0272 |
| AveragePool3D(2, 1) | 0.57 | 0.69 | 0.77 | 582.2272 |
| AveragePool3D(2, 2) | 0.75 | 0.82 | 0.46 | 14.1437 |
| AveragePool3D(3, 1) | 0.61 | 0.68 | 0.62 | 60.6249 |
| AveragePool3D(3, 2) | **0.76** | **0.84** | 0.38 | 0.1450 |
| AveragePool3D(4, 1) | 0.70 | 0.75 | **0.31** | 0.5721 |

**Table S2.** The OOD detection hyperparameter search for PCA. Bold text denotes the best performance. PCA(n) represents PCA being performed with n components. For AUROC and AUPR, higher is better. For FPR75, lower is better. Computation time is the time it takes the calculate the inverse of the covariance matrix in the Mahalanobis distance calculation in seconds.

| Experiment | AUROC | AUPR | FPR75 | Computation Time |
|---|---|---|---|---|
| PCA(2) | 0.90 | 0.93 | **0.07** | 0.0001 |
| PCA(4) | 0.70 | 0.66 | 0.38 | 0.0002 |
| PCA(8) | 0.73 | 0.74 | 0.46 | 0.0003 |
| PCA(16) | 0.87 | 0.87 | 0.23 | 0.0004 |
| PCA(32) | 0.86 | 0.88 | 0.23 | 0.0005 |
| PCA(64) | 0.82 | 0.85 | 0.23 | 0.0005 |
| PCA(128) | 0.89 | 0.93 | 0.15 | 0.0011 |
| PCA(256) | **0.93** | **0.94** | 0.14 | 0.0106 |

**Table S3.** The OOD detection hyperparameter search for UMAP. Bold text denotes the best performance. UMAP(n) represents UMAP being performed with n components. For AUROC and AUPR, higher is better. For FPR75, lower is better. Computation time is the time it takes the calculate the inverse of the covariance matrix in the Mahalanobis distance calculation in seconds. All results are the average (±SD), n=10.

| Experiment | AUROC | AUPR | FPR75 | Computation Time |
|---|---|---|---|---|
| UMAP(2) | **0.79 (±0.05)** | **0.85 (±0.04)** | **0.36 (±0.13)** | 0.0002 (±0.0000) |
| UMAP(4) | 0.76 (±0.06) | 0.83 (±0.04) | 0.46 (±0.14) | 0.0002 (±0.0000) |
| UMAP(8) | 0.74 (±0.06) | 0.83 (±0.04) | 0.43 (±0.20) | 0.0002 (±0.0001) |
| UMAP(16) | 0.71 (±0.06) | 0.80 (±0.04) | 0.51 (±0.17) | 0.0001 (±0.0001) |
| UMAP(32) | 0.71 (±0.04) | 0.80 (±0.03) | 0.42 (±0.10) | 0.0002 (±0.0001) |
| UMAP(64) | 0.69 (±0.05) | 0.79 (±0.03) | 0.46 (±0.14) | 0.0003 (±0.0000) |
| UMAP(128) | 0.64 (±0.05) | 0.76 (±0.02) | 0.64 (±0.15) | 0.1372 (±0.1390) |
| UMAP(256) | 0.62 (±0.06) | 0.75 (±0.03) | 0.69 (±0.15) | 0.3558 (±0.2820) |